\documentclass[letterpaper, 10 pt, conference]{ieeeconf}  

\IEEEoverridecommandlockouts
\usepackage{amsthm}

\usepackage{graphicx} 
\usepackage{amsmath} 
\usepackage{amssymb}  
\usepackage{mathtools} 
\usepackage{subcaption} 
\usepackage{comment}
\usepackage{cite} 
\usepackage[font=footnotesize]{caption} 
\usepackage[ruled,vlined,linesnumbered]{algorithm2e} 
\usepackage{bm} 
\usepackage{xcolor} 

\usepackage{hyperref} 

\title{\LARGE \bf
Signal Temporal Logic-Guided Model Predictive Control for Robust Bipedal Locomotion Resilient to Runtime External Perturbations
}

\author{Zhaoyuan Gu, Rongming Guo, William Yates, Yipu Chen, and Ye Zhao
\thanks{The authors are with the Laboratory for Intelligent Decision and Autonomous Robots, Woodruff School of Mechanical Engineering, Georgia Institute of Technology. {\tt\footnotesize \{zgu78, yezhao\}@gatech.edu}}
\thanks{This work was funded by the NSF grant \# IIS-1924978, \# CMMI-2144309, and Office of Naval Research AWD\# N00014-23-1-2223.}
}

\begin{document}

\bstctlcite{IEEEexample:BSTcontrol} 

\maketitle
\thispagestyle{empty}
\pagestyle{empty}

\section{Introduction}
This study investigates formal-method-based trajectory optimization (TO) \cite{Lin_Smooth} for bipedal locomotion, focusing on scenarios where the robot encounters external perturbations at unforeseen times \cite{RMP, MIT_CBF, IHMC_cross}. Our key research question centers around the assurance of task specification correctness \cite{shamsah2023integrated, zhao2018reactive, LTL_Nav_Kulgod, LTL_Nav_Warnke} and the maximization of specification robustness for a bipedal robot in the presence of external perturbations. 

\begin{figure}[h]
\centerline{\includegraphics[width=.48\textwidth]{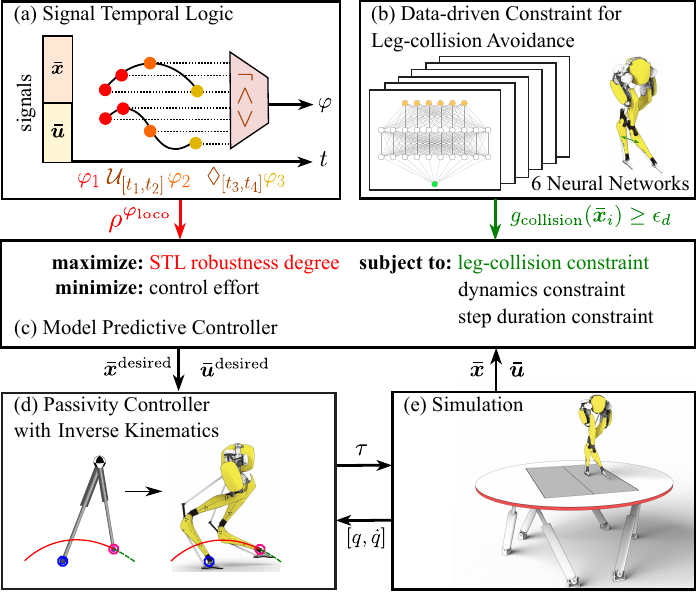}}
\caption{
Block diagram of the proposed framework. (a) The STL specification $\varphi$ specifies the signals to achieve a stable walking task. (b) A set of data-driven kinematic constraints enforce the leg self-collision avoidance. (c) The MPC solves the control synthesis problem. (d) A passivity-based controller tracks the synthesized plan. (e) The walking experiment with external perturbation. $\boldsymbol{\bar{x}}$ and $\boldsymbol{\bar{u}}$ are the state and control of the reduced-order model. $\tau$ and $q$ represent the joint torque and joint position, respectively.}
\label{fig:framework}
\vspace{-0.2in}
\end{figure}

Our contribution includes the design of an optimization-based task and motion planning framework (illustrated in Fig.~\ref{fig:framework}) that generates optimal control sequences with formal guarantees of external perturbation recovery. As a core component of the framework, a model predictive controller (MPC) encodes signal temporal logic (STL)-based task specifications as a cost function. In particular, we investigate challenging scenarios where the robot is subjected to lateral perturbations that increase the risk of failure due to leg self-collision. To address this, we synthesize agile and safe crossed-leg maneuvers to enhance locomotion stability.

This work marks the first study to incorporate formal guarantees offered by STL \cite{Belta_STL_review} into a TO for perturbation recovery of bipedal locomotion.
We demonstrate the efficacy of the framework via perturbation experiments in simulations. 

This work is significantly distinct from our previous study \cite{Gu_push} in the following aspects.
First, the high-level task planner in \cite{Gu_push} designs an abstraction-based decision-maker using linear temporal logic (LTL) while our study employs an optimization-based MPC using STL to allow dense-time signals. This property eliminates the previously existing mismatch between the high-level discrete abstraction and low-level continuous robot dynamics. 
Second, the STL formulation eliminates the need for environmental perturbation modeling. This allows for more forceful perturbations that would lead to undefined abstraction in the LTL-based method.
Overall, the STL-based task planner better suits locomotion recovery from unknown perturbations that require frequent reactive planning.

\section{Signal temporal logic}
Signal temporal logic (STL) is used to formulate a locomotion specification $\varphi_{\rm loco}$, which is composed of a foot location bound $\varphi_{\rm foot}$ and a stability specification $\varphi_{\rm stable}$: $$\varphi_{\rm loco} = (\square \varphi_{\rm foot}) \wedge \varphi_{\rm stable}$$ 
where $\varphi_{\rm foot}$ restricts the foot placement to \textit{always} reside within the treadmill's surface area while the stability specification $\varphi_{\rm stable}$ indicates that the robot \textit{eventually} recovers to a periodic gait within a finite time horizon.

Periodic walking is described using two subformulas: the keyframe specification $\varphi_{\rm keyframe}$ and the Riemannian specification $\varphi_{\rm riem}$. $\varphi_{\rm keyframe}$ signals the locomotion apex state when the center-of-mass (CoM) passes the stance foot in the sagittal direction. 
$\varphi_{\rm riem}$ monitors whether or not the state signal is within a state-space robust region, named Riemannian region \cite{Zhao2017IJRR}. Both $\varphi_{\rm keyframe}$ and $\varphi_{\rm riem}$ must be true at the same time to guarantee that the keyframe state is on a stable gait.

Another feature of STL is that it admits \textit{quantitative semantics}, which denote the \textit{robustness degree} \cite{MTL_Papas} of specification satisfaction \cite{Belta_STL_review}. The satisfaction of the specification $\varphi_{\rm loco}$ is equivalent to the STL robustness being positive: $(\boldsymbol{y},t) \models \varphi_{\rm loco} \Leftrightarrow \rho^{\varphi_{\rm loco}}(\boldsymbol{y},t) \geq 0$. Therefore, to ensure the \textit{correct-by-construction} property, we encode the STL robustness as a constraint in the TO, thereby guaranteeing the solution to satisfy $\varphi_{\rm loco}$. Additionally, the robustness of a walking step is measured as the minimum Riemannian distance between the CoM keyframe state and the bounds of the Riemannian region. Using this measure, we maximize the robustness degree of locomotion as an objective of the proposed MPC in the next section.

The main contribution of this specification design lies in its effectiveness and simplicity. The synthesis problem based on this specification is solved rapidly for online reactive planning and generates versatile locomotion behaviors such as crossed-leg maneuvers. 

\section{Model predictive control for push recovery}

The model predictive controller (MPC) is designed to solve a trajectory optimization problem, synthesizing a sequence of state and control signals that satisfy the locomotion specification $\varphi_{\rm loco}$, the system dynamics, and self-collision avoidance constraints. 

This problem is formulated as a nonlinear program (NLP) as shown in Fig.~\ref{fig:framework}(c).
The MPC utilizes the robustness objective from STL, aiming to maximize the robustness degree of the task specification $\rho^{\varphi_{\rm loco}}$ while simultaneously minimizing the control cost to ensure a smooth trajectory. 
The dynamics constraints are hybrid, including an enhanced linear inverted pendulum model (LIPM) composed of the CoM and swing foot states, along with a \textit{reset map} that models the ground-foot contacts. To improve robust locomotion recovery, the MPC modifies the walking-step durations \cite{step_time_adaptation, Griffin_step_up}. Moreover, a set of data-driven kinematic constraints are learned to prevent leg self-collision. These constraints are multi-layer perceptrons that represent accurate collision detections based on enhanced LIPM states.

At the low level, a passivity-based whole-body controller tracks the centroidal and swing-leg states. The MPC operates on an Intel i7-1260P CPU at $30$ Hz, while the controller runs on a separate computer at $2$ kHz. We demonstrate the efficacy of our framework in the following section. 

\begin{figure}[t!]
\centerline{\includegraphics[width=.48\textwidth]{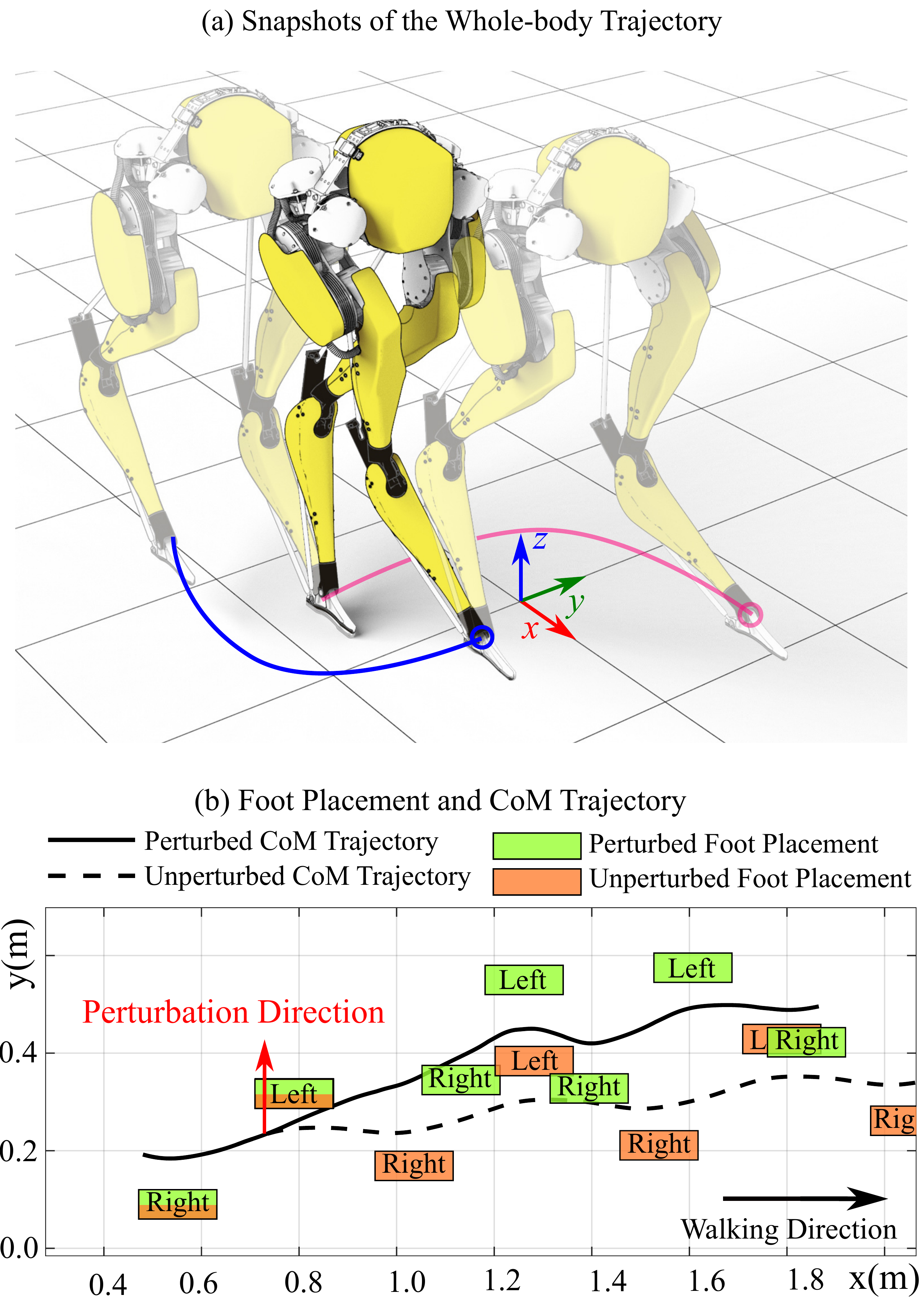}}
\caption{
(a) Snapshots of the whole body trajectory during a cross-leg maneuver. (b) The CoM and foot placement plan executed by the robot during normal and perturbed walking processes.}
\label{fig:collision_avoid}
\vspace{-0.15in}
\end{figure}

\section{Results}

A motion plan generated by our STL-MPC is portrayed in Fig.~\ref{fig:collision_avoid} to showcase its effectiveness towards a particular push recovery setup in simulation, where an adversarial perturbation brings the robot's CoM over its stance foot and leaves leg crossover as the only viable strategy for recovery. As shown in Fig.~\ref{fig:collision_avoid}(a), the robot adeptly maneuvers its swing leg around its stance leg to actively avoid self-collision as the legs cross and uncross. Fig.~\ref{fig:collision_avoid}(b) illustrates the simulated CoM trajectory and foot placement plan.

\begin{figure}[ht!]
\centerline{\includegraphics[width=.46\textwidth]{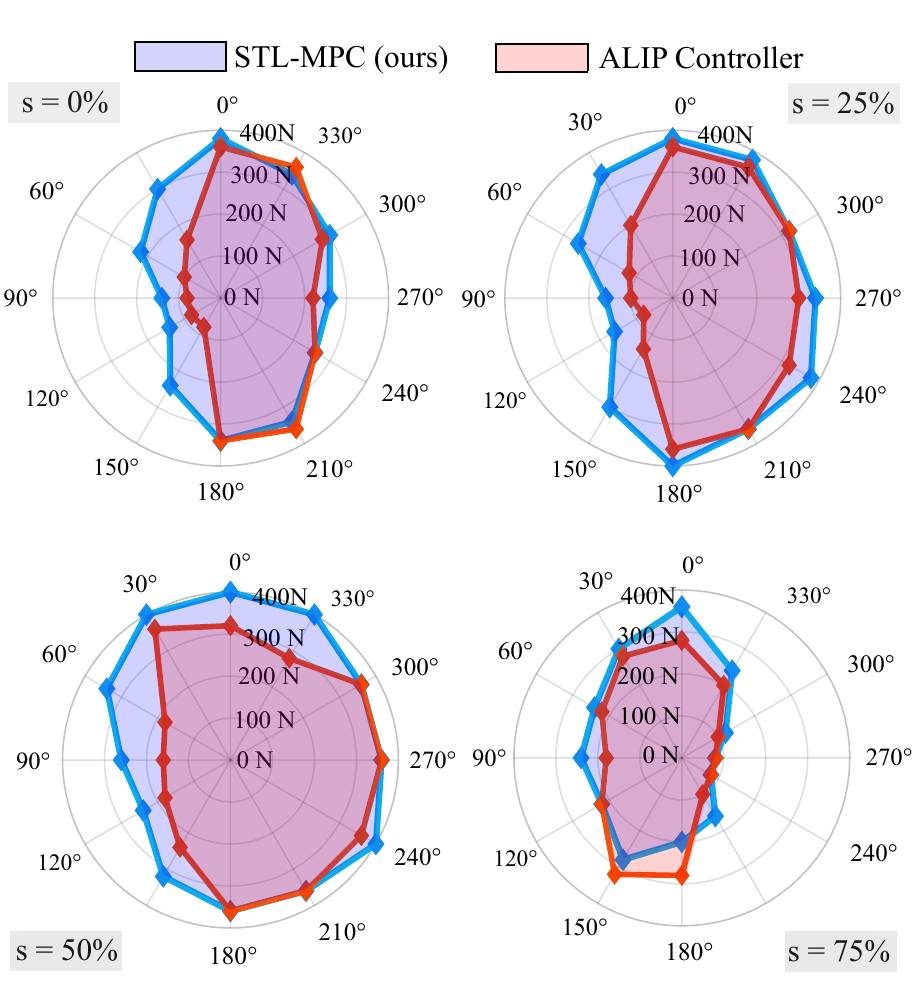}}
\caption{
The maximum force exerted on the pelvis from which the robot can safely recover within two steps in all 12 directions. The perturbations happen at different phases $s$ during a left leg stance. Values on the left half result in crossed-leg maneuvers, and values on the right half correspond to wide-step recoveries.}
\label{fig:spider}
\vspace{-0.25in}
\end{figure}

We further verify the robustness of the entire framework through perturbation experiments in a high-fidelity simulation. The bipedal robot Cassie is systematically perturbed with varied magnitudes, directions, and timings. The STL-MPC has a horizon of three walking steps and is solved online at $30$ Hz. A low-level passivity-based controller operates at $2$ kHz. 
In Fig.~\ref{fig:spider}, we compare the maximum impulse the STL-MPC can withstand to that of a baseline foot placement controller \cite{MomentumController}. The STL-MPC demonstrates superior perturbation recovery performance across the vast majority of directions and phases, as reflected by the blue region encompassing the red region.


\newpage
\bibliographystyle{ieeetran}
\bibliography{references}

\end{document}